\documentclass[11pt,a4paper]{article}
\usepackage[hyperref]{acl2021}
\usepackage{times}
\usepackage{graphicx}
\usepackage{latexsym}
\usepackage{breqn}
\usepackage{amsmath}
\usepackage{amssymb}
\usepackage{amsfonts}
\usepackage{amstext}
\usepackage{amsthm}

\usepackage{microtype}

\aclfinalcopy % Uncomment this line for the final submission
\title{Cisco at SemEval-2021 Task 5: What’s Toxic?: Leveraging Transformers
for Multiple Toxic Span Extraction from Online Comments}

\author{Sreyan Ghosh \\
  Cisco Systems, Bangalore, India \\
  MIDAS Lab, IIIT-Delhi, India \\
  \texttt{sreyghos@cisco.com} \\\And
  Sonal Kumar \\
  Cisco Systems, Bangalore, India \\
  \texttt{sonalkum@cisco.com} \\}

\date{}

\begin{document}
\maketitle
\begin{abstract}
Social network platforms are generally used to share positive, constructive, and insightful content. However, in recent times, people often get exposed to objectionable content like threat, identity attacks, hate speech, insults, obscene texts, offensive remarks or bullying. Existing work on toxic speech detection focuses on binary classification or on differentiating toxic speech among a small set of categories. This paper describes the system proposed by team Cisco for SemEval-2021 Task 5: Toxic Spans Detection, the first shared task focusing on detecting the spans in the text that attribute to its toxicity, in English language. We approach this problem primarily in two ways: a sequence tagging approach and a dependency parsing approach. In our sequence tagging approach we tag each token in a sentence under a particular tagging scheme. Our best performing architecture in this approach also proved to be our best performing architecture overall with an \emph{F\textsubscript{1}} score of \textbf{0.6922}, thereby placing us $7^{th}$ on the final evaluation phase leaderboard. We also explore a dependency parsing approach where we extract spans from the input sentence under the supervision of target span boundaries and rank our spans using a biaffine model. Finally, we also provide a detailed analysis of our results and model performance in our paper.
\end{abstract}

\section{Introduction}
It only takes one toxic comment to sour an online discussion. The threat of abuse and harassment online leads many people to stop expressing themselves and give up on seeking different opinions. Toxic content is ubiquitous in social media platforms like Twitter, Facebook, Reddit, the increase of which is a major cultural threat and has already lead to a crime against minorities \citep{williams2020hate}. Toxic text in online social media varies depending on targeted groups (e.g. women, LGBT, gay, African, immigrants) or the context (e.g. pro-trump discussion or the \#metoo movement). Toxic Text online has often been broadly classified by researchers into different categories like hate, offense, hostility, aggression, identity attacks, and cyberbullying. Though the use of various terms for equivalent tasks makes them incomparable at times \citep{fortuna2020toxic}, toxic speech or spans in this particular task, SemEval-2021 Task 5 \citep{pav2020semeval}, has been considered as a super-set of all the above sub-types.
\begin{figure}[ht]
\centering
\includegraphics[width=0.43\textwidth]{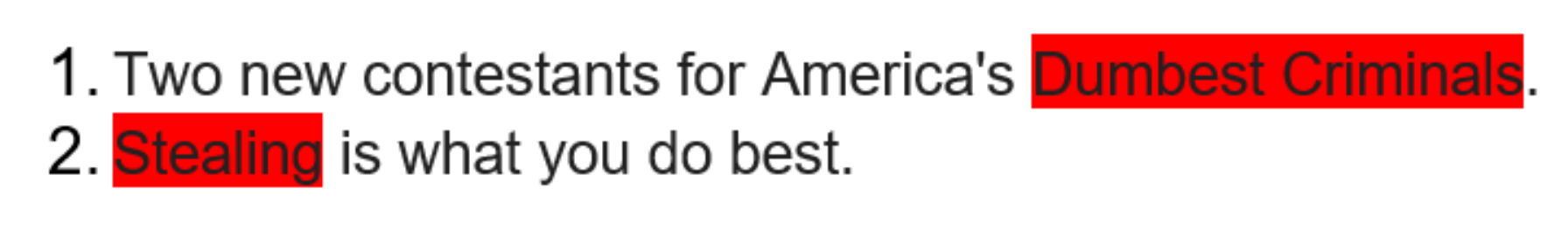}
\caption{Toxic spans in sentences}
\label{toxic_spans_in_sentences}
\end{figure}

While a lot of models have claimed to achieve state-of-the-art results on various datasets, it has been observed that most models fail to generalize \citep{arango2019hate,grondahl2018all}. The models tend to classify comments as toxic that have a reference to certain commonly-attacked entities (e.g. gay, black, Muslim, immigrants) without the comment having any intention to be toxic \citep{dixon2018measuring,borkan2019nuanced}. A large vocabulary of certain trigger terms leads to a biased prediction by the models \citep{sap2019risk,davidson2017automated}. Thus, it has become increasingly important in recent times to determine parts of the text that attribute to the toxic nature of the sentence, for both automated and semi-automated content moderation on social media platforms, primarily for the purpose of helping human moderators deal with lengthy comments and also provide them attributions for better explainability on the toxic nature of the post. This in turn would aid in better handling of unintended bias in toxic text classification. SemEval-2021 Task 5: Toxic Spans Detection focuses on exactly this problem of detecting toxic spans from sentences already classified as toxic on a post-level.

In this paper, we approach the problem of multiple non-contiguous toxic span extraction from texts both as a \emph{sequence tagging task} and as a standard \emph{span extraction task} resembling the generic approach and architecture adopted for single-span Reading Comprehension (RC) task. For our sequence tagging approach, we predict for each token, whether it is a part of the span. For our second approach, we predict and compute a couple of scores for each token, corresponding to whether that token is the start or end of the span. In addition to this, we deploy a biaffine model to score start and end indices, thus adopting the methodology for multiple non-contiguous span extraction. 

\section{Literature}

Previous work on automated toxic text detection, and its various sub-types, focuses on developing classifiers that can flag toxic content with a high degree of accuracy on datasets curated from various social media platforms in English\cite{carta2019supervised,saeed2018overlapping,vaidya2020empirical}, other foreign languages \citep{zhang2018detecting,mishra2018author,qian2019benchmark,davidson2017automated,kamal2021hostility, leite2020toxic} including code-switched text \citep{mathur2018did,mathur2018detecting,kapoor2019mind} and multilingual text \citep{zampieri2019semeval}. This topic has also evidenced a number of workshops \citep{kumar2018benchmarking} and competitions \citep{zampieri2019semeval,zampieri2020semeval,basile-etal-2019-semeval, mandl2019overview}.

Recent work shows transformer based architectures like BERT \citep{devlin-etal-2019-bert} have been performing well on the task of offensive language classification \citep{liu2019nuli,safaya2020kuisail,dai2020kungfupanda}.
Transformer based architectures have also produced state-of-the-art performance on sequence tagging tasks like \emph{Named Entity Recognition (NER)} \citep{yamada2020luke,devlin-etal-2019-bert,yang2019xlnet} \emph{span extraction} \citep{eberts2019span,joshi2020spanbert} and \emph{QA tasks} \citep{devlin-etal-2019-bert,yang2019xlnet,Lan2020ALBERT:}. Multiple span extraction from texts has been explored both as a \emph{sequence tagging task} \citep{patil2020bpgc,segal2019simple} and as span extraction as in RC tasks\citep{hu2019open,yu2020named}.

Very recently HateXplain \citep{mathew2020hatexplain} proposed a benchmark dataset for explainable hate speech detection using the concept of rationales. Attempts have also been made to handle identity bias in toxic text classification \citep{vaidya2020empirical} and also to make robust toxic text classifiers which help adversaries not bypass toxic filters \citep{kurita2019towards}.

\section{Methodology}

For our sequence tagging approach, we explore two tagging schemes. First, the well known \emph{BIO} tagging scheme, where \emph{B} indicates the first token of an output span, \emph{I} indicates the subsequent tokens and \emph{O} denotes  the tokens that are not part of the output span. Additionally, we also try a simpler \emph{IO} tagging scheme, where words which are part of a span are tagged as \emph{I} or \emph{O} otherwise. Formally, given an input sentence \textbf{x} = (\emph{x\textsubscript{1}},...,\emph{x\textsubscript{n}}), of length n,and a tagging scheme with $|\emph{S}|$ tags ($|\emph{S}|$ = 3 for BIO and $|\emph{S}|$ = 2 for IO), for each of \emph{n} tokens the probability for the tag of the \emph{i}-th token is 
\begin{equation}
\textbf{p}_{i}=softmax(f(\textbf{h}_{i}))
\end{equation}
where \textbf{p} $\in\mathbb{R}^{m \times|S|}$ and $f$ is a paremeterized function with $|S|$.

Our other approach is based on the standard single-span extraction architecture widely used for RC Tasks. With this approach, we extract toxic spans from sentences under the supervision of target span boundaries, but with an added biaffine model for scoring the multiple toxic spans instead of simply taking top k spans based on the start and end probabilities, thus giving our model a global view of the input. The main advantage of this approach is that the extractive search space can be reduced linearly with the sentence length, which is far less than the sequence tagging method. Given an input sentence \textbf{x} = ($x_1$,...,$x_n$), of length n, we predict a target list \textbf{T} = ($t_1$,...,$t_m$)  where the number of targets is $m$ and each target $t_i$ is annotated with its start position $s_i$, its end position $e_i$ and the class that span belongs to (only one in our case, $toxic$). 

However, to adapt to the problem of extracting multiple spans from the sentence, instead of taking the top k spans based on the start and end probabilities, we apply a biaffine model \citep{DBLP:journals/corr/DozatM16} to score all the spans with the constraint \emph{s\textsubscript{i}} $\leq$ \emph{e\textsubscript{i}}. Post this we rank all the spans in descending order and choose every span as long it does not clash with higher-ranked spans.

\section{Dataset}

The dataset provided to us by the organizers of the workshop consisted of a random subset of 10,000 posts from the publicly available Civil Comments Dataset, from a set of 30,000 posts originally annotated as toxic (or severely toxic) on post-level annotations, manually annotated by 3 crowd-raters per post for toxic spans. The final character offsets were obtained by retaining the offsets with a probability of more than 50\%, computed as a fraction of raters who annotated the character offsets as toxic. Basic statistics about the dataset can be found in Table 1.

\begin{table}[ht]
    \centering
    \begin{tabular}{l l l}
    \hline
         & Sentences & Spans  \\
         \hline
          Train & 7939 & 10298 \\
          Dev & 690 & 903\\
          Test & 2000 & 1850\\
          \hline
    \end{tabular}
    \caption{Number of sentences and spans}
    \label{tab:num_of_sent}
\end{table}

Additionally, we provide a quick look into the length-wise distribution of spans across the train, development, and test set in Table 2. As we observe, the majority of the spans are just a single word in length and mostly comprise of the most commonly used cuss words in the \emph{English} language. In our Results Analysis section, we show how this metric stands important for training and evaluating our systems and for the future development of toxic span extraction datasets.

\begin{table}[ht]
    \centering
    \begin{tabular}{l l l l}
    \hline
         & Train & Dev & Test \\
         \hline
        1 & 7897 & 687 & 1650 \\
        2-4 & 1617  & 153 & 174\\
        $>=$5 & 784 & 63 & 26\\
        \hline
    \end{tabular}
    \caption{Length-wise segregation of the number of non-contiguous spans}
    \label{tab:length_wise}
\end{table}

\section{Evaluation Metric}
To evaluate the performance of our systems we employ \emph{F\textsubscript{1}} as used by \citet{da2019fine}. Let system \emph{A} return a set \emph{S\textsuperscript{t}\textsubscript{A}} of character offsets, for parts of the post found to be toxic. Let \emph{S\textsuperscript{t}\textsubscript{G}} be the character offsets of the ground truth annotations of post \emph{t}. We calculate \emph{F\textsubscript{1}} score of \emph{S\textsuperscript{t}\textsubscript{A}} w.r.t \emph{S\textsuperscript{t}\textsubscript{G}} as follows where $|.|$ denotes set cardinality.

\begin{equation}
P^{t}\left(A, G\right)=\frac{\left|S_{A}^{t} \cap S_{G}^{t}\right|}{\left|S_{A}^{t}\right|}
\end{equation}

\begin{equation}
R^{t}\left(A, G\right)=\frac{\left|S_{A}^{t} \cap S_{G}^{t}\right|}{\left|S_{G}^{t}\right|}
\end{equation}

\begin{equation}
F_{1}^{t}\left(A, G\right)=\frac{2 \cdot P^{t}\left(A, G\right) \cdot R^{t}\left(A, G\right)}{P^{t}\left(A, G\right)+R^{t}\left(A, G\right)}
\end{equation}

If predicted span i.e \emph{S\textsuperscript{t}\textsubscript{A}} is empty for a post \emph{t} then we set \emph{F\textsuperscript{t}\textsubscript{1}(A,G)} = 1 if the gold truth i.e \emph{S\textsuperscript{t}\textsubscript{G}} is also empty, else if \emph{S\textsuperscript{t}\textsubscript{G}} is empty and \emph{S\textsuperscript{t}\textsubscript{A}} is not empty then we set \emph{F\textsuperscript{t}\textsubscript{1}(A,G)} = 0.

\section{System Description}
\subsection{Sequence Tagging Approach}

For our sequence tagging approach we employ the commonly used BiLSTM-CRF architecture \citep{huang2015bidirectional} used predominately in many sequence tagging problems, but with added contextual word embeddings for each word using transformer and character-based word embeddings. We experiment with a total of 5 transformer architectures, namely BERT \citep{devlin-etal-2019-bert}, XLNet \citep{yang2019xlnet}, RoBERTa \citep{liu2019roberta}, ALBERT \citep{Lan2020ALBERT:} and SpanBERT \citep{joshi2020spanbert}. For all of the above mentioned transformer architectures, the \emph{large} variant of the transformer was used except ALBERT for which we use its \emph{xlarge-v2} variant. First, the tokenized word input is passed through the transformer architecture and the output of the last 4 encoder layers is concatenated to obtain the final contextualized word embedding \emph{E\textsubscript{T}} for each word in the sentence. Additionally, we also pass each character in a word through a character-level BiLSTM network, to obtain character-based word embeddings for the word \emph{E\textsubscript{C}} as used by \citet{lample-etal-2016-neural}. Finally, both these word embeddings, \emph{E\textsubscript{T}} and \emph{E\textsubscript{C}}, for each word are concatenated and passed through a BiLSTM layer followed by a CRF layer to obtain the best probable tag for each word in the sentence.
\begin{figure}[ht]
\centering
\includegraphics[width=0.44\textwidth]{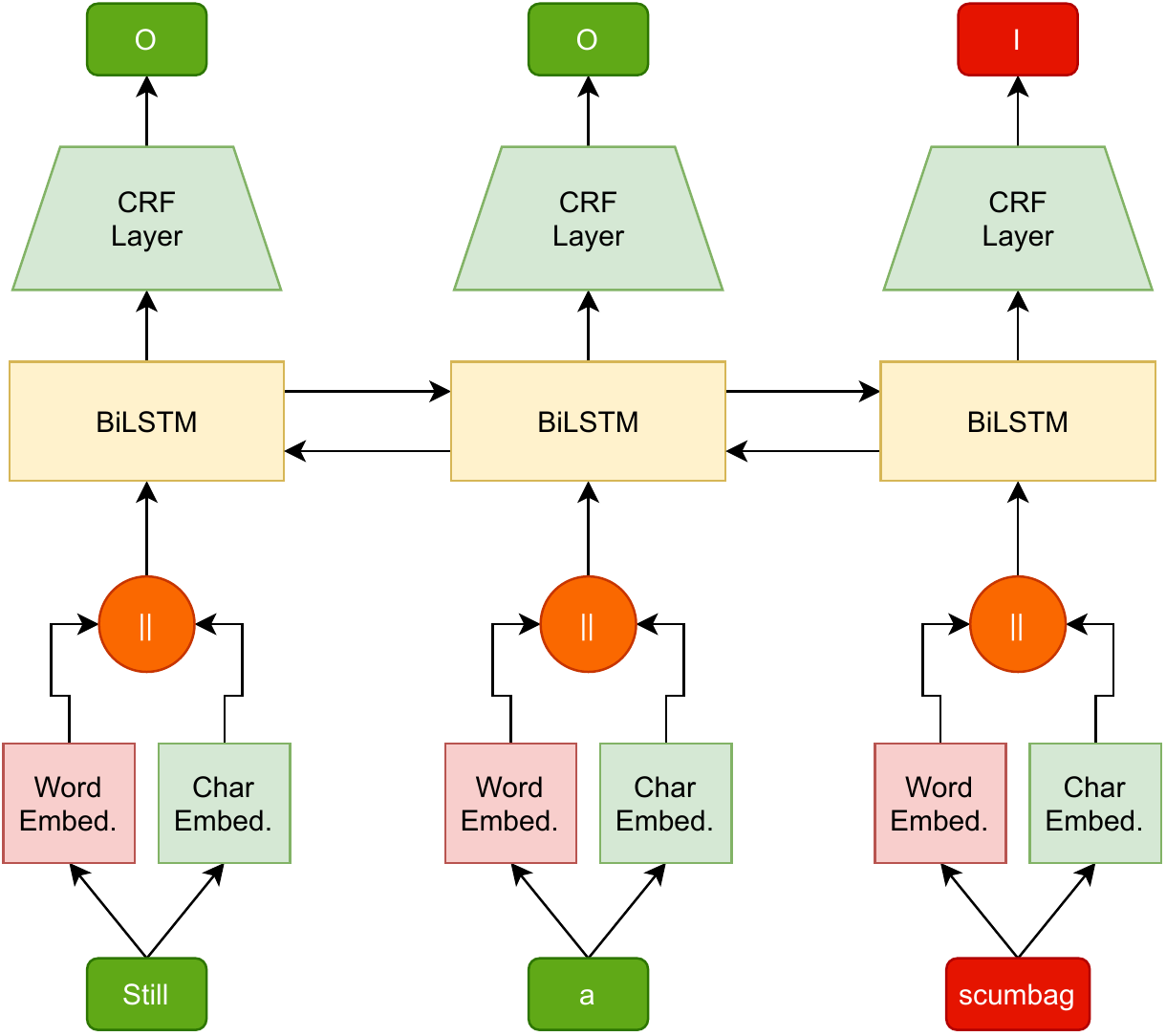}
\caption{Sequence Tagger Model}
\label{Sequence_Tagger_Model}
\end{figure}

\subsection{Dependency Parsing Approach}
For our dependency parsing approach, we employ a similar approach as proposed by \citet{yu2020named}, using a biaffine classifier to score our spans post-extraction. This methodology fits best to our purpose of \emph{multiple} toxic span extraction from sentences compared to span extraction systems in general RC tasks which are capable of extracting just a single span from a sentence \citep{yangquestion}. 
For each word first we extract it's BERT, FasText and character-based word embeddings. We used BERT\textsubscript{Large} for all our experiments and used the recipe followed by \citet{kantor-globerson-2019-coreference} to extract contextual embeddings for each token. After concatenating both the word embeddings and character embeddings for each word, we feed the output to a BiLSTM layer. We then apply two separate FFNNs to the output word representations $x$ to create different representations ($h_s$ / $h_e$) for the start/end of the spans. These representations are then passed through a biaffine model for scoring all possible spans ($s_i$,$e_i$), where $s_i$ and $e_i$ are start and end indices of the span, under the constraint $s_i$ $\leq$ $e_i$ (the start of the span is before its end) by creating a $ l \times l \times c$ scoring tensor $r_m$, where $l$ is the length of the sentence and $c$ is the number of NER categories + 1(for non-entity). We compute the score for a span $i$ by:
\begin{equation}
h_{s}(i)=\mathrm{FFNN}_{s}\left(x_{s_{i}}\right)
\end{equation}

\begin{equation}
h_{e}(i)=\mathrm{FFNN}_{e}\left(x_{e_{i}}\right)
\end{equation}

\begin{equation}
\begin{split}
r_{m}(i) = & h_{s}(i)^{\top} \mathrm{U}_{m} h_{e}(i) \\
         &  + W_{m}\left(h_{s}(i) \oplus h_{e}(i)\right)+b_{m}
\end{split}
\end{equation}

We finally assign each span a category $y{\prime}$ based on 

\begin{equation}
y^{\prime}(i)=\arg \max r_{m}(i)
\end{equation}

%Post this, we rank each span that has a category other than non-entity and considers all spans for our final prediction as long its indices do not clash with or isn't inside a higher-ranked span.
Post this, we rank each span that has a category other than non-entity and consider all the spans for our final prediction as long as it does not clash with higher ranked spans with an additional constraint, whereby, an entity containing or is inside an entity ranked before it will not be selected.

\begin{figure}[ht]
\centering
\includegraphics[width=0.43\textwidth]{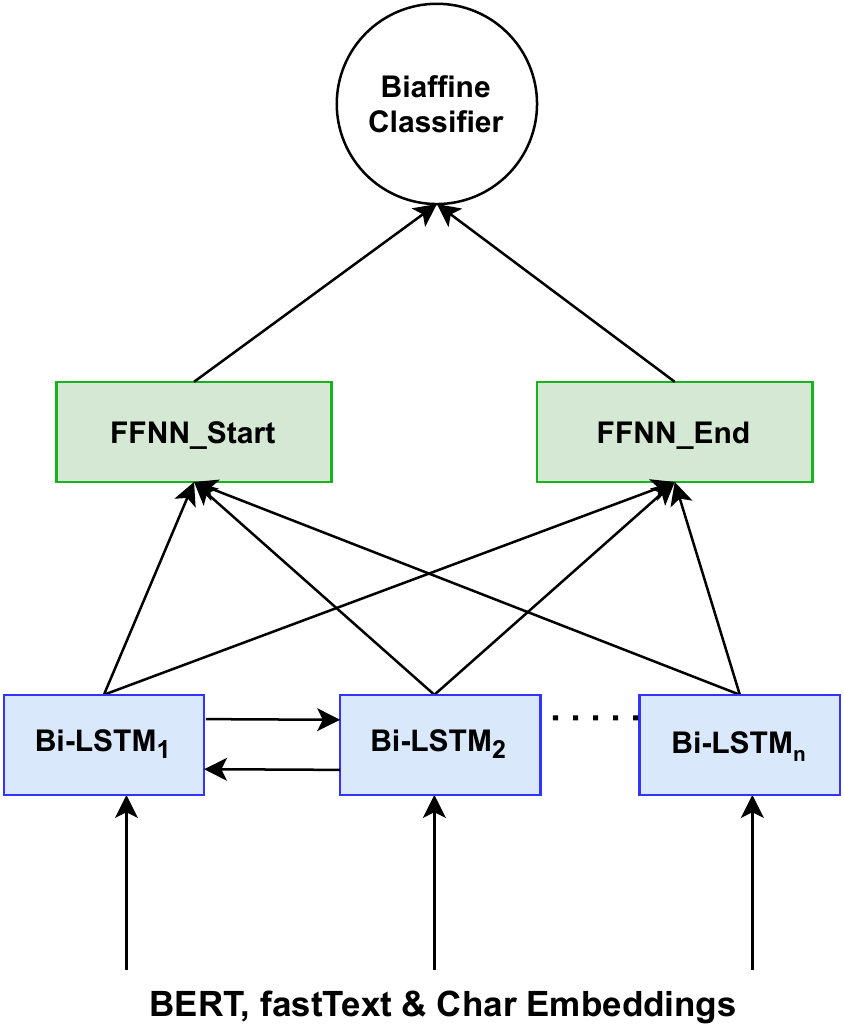}
\caption{Biaffine Model}
\label{Biaffine_Model}
\end{figure}

\section{Experimental Setup}

Data was originally provided to us in the form of sentences and the corresponding character offsets for the toxic spans of the sentence. Before converting the character offsets to our required format for our respective approaches, we apply some basic text pre-processing to all our sentences. First, we normalize all the sentences by converting all white-space characters to spaces. Second, we split all punctuation characters from both sides of a word and also break abbreviated words. These pre-processing steps help improve the \emph{F\textsubscript{1}} score of both our approaches as shown in Table 6. Post these pre-processing steps, we formulate our targets for both our approaches. For our sequence tagging approach, we tag each word in the sentence with its corresponding tag based on the tagging scheme we follow, \emph{BIO} or \emph{IO}. For our span extraction approach, we convert the sequence of character offsets into its corresponding word-level start and end indices for each span. In Fig. 4, we provide a pictorial representation of the above mentioned procedures we follow for data preparation for both our approaches.

%In Fig.3 we provide a pictorial representation of the target formulation for both our approaches from the raw data, after the pre-processing steps.

\begin{figure}[ht]
\centering
\includegraphics[width=0.48\textwidth]{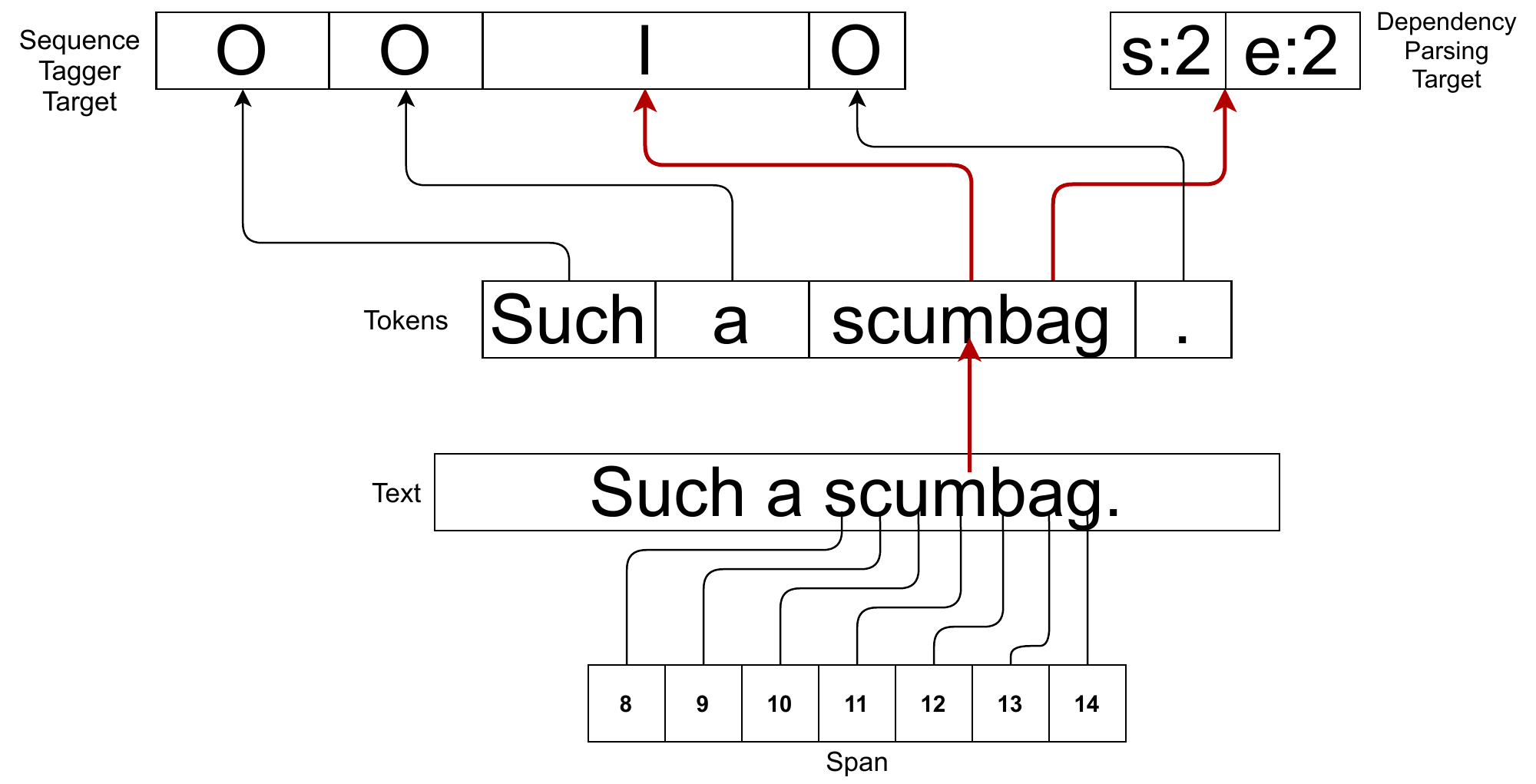}
\caption{Data Preparation}
\label{data_annot}
\end{figure}

We use PyTorch\footnote{https://pytorch.org/} Framework for building our Deep Learning models
along with the Transformer implementations, pre-trained
models and, specific tokenizers in the HuggingFace\footnote{http://huggingface.co/} library.

We mention the major hyperparameters of our best-performing systems experimental setting for our dependency parsing approach and span extraction approach in Tables 3 and 4 respectively.

\begin{table}[ht]
    \centering
    \begin{tabular}{l l}
    \hline
        Parameter & Value \\
    \hline
        BiLSTM size & 256 \\
        BiLSTM layer & 1 \\
        BiLSTM dropout & 0 \\
        Transformer size & 1024 \\
        Transformer encoder layers & last 4 \\
        Char BiLSTM Hidden Size & 25 \\
        Char BiLSTM layers & 1 \\
        Optimiser & Adam \\
        Learning rate & [1e-3,1.56e-4]  \\
    \hline
    \end{tabular}
    \caption{Major hyperparameters of Sequence Tagger model}
    \label{tab:paremeter_value}
\end{table}

\begin{table}[ht]
    \centering
    \begin{tabular}{l l}
    \hline
        Parameter & Value \\
    \hline
        BiLSTM size & 200 \\
        BiLSTM layer & 3 \\
        BiLSTM dropout & 0.4 \\
        FFNN size & 150 \\
        FFNN dropout & 0.2 \\
        BERT size & 1024 \\
        BERT encoder layers & last 4 \\
        fastText embedding size & 300 \\
        Char CNN size & 50 \\
        Char CNN filter width & [3,4,5] \\
        Embeddings dropout & 0.5 \\
        Optimiser & Adam \\
        Learning rate & 1e-3 \\
    \hline
    \end{tabular}
    \caption{Major hyperparameters of Dependancy Parsing model}
    \label{tab:param_val}
\end{table}

We train all our sequence tagging models with stochastic gradient descent in batched mode with a batch size of 8. In the training phase, we keep all layers in our model, including all the transformer layers trainable. We start training our model at a learning rate of 0.01, with a minimum threshold limit of 0.0001, and half the learning rate after every 4 consecutive epochs of no improvement in the \emph{F\textsubscript{1}} score of the development set. We train our model to a maximum of 100 epochs or 4 consecutive epochs of no improvement at our minimum learning rate.

We train our our model for dependency parsing approach with Adam optimizer in batched mode with a batch size of 32 and a learning rate of 0.0001 for a maximum of 40,000 steps. With this approach too, we keep all layers trainable in the training phase except the BERT Transformer layers. Pre-trained BERT and fastText embeddings were just used to extract context-dependent and independent embeddings respectively and BERT was \emph{not fine-tuned} in the training phase.

The training was performed on 1 NVIDIA Titan X
GPU. Our code is available on Github\footnote{https://github.com/Sreyan88/SemEval-2021-Toxic-Spans-Detection}.

\section{Results}

In Table 5 we present \emph{F\textsubscript{1}} scores for all our systems trained for both our sequence tagging and span extraction approaches. For our sequence tagging approach, we divide our results according to the transformer architecture and tagging scheme used for that experiment.

\begin{table}[ht]
\centering
\begin{tabular}{l l c c c}
\hline
Model & Scheme & Test & Dev \\
\hline
XLNet & IO & \textbf{0.6922} & 0.6945 \\
XLNet & BIO & 0.6653 & 0.6683 \\
spanBERT & IO & 0.6777 & 0.6744 \\
spanBERT & BIO & 0.6887 & 0.6730 \\
RoBERTa & IO & 0.6647 & \textbf{0.6967} \\
RoBERTa & BIO & 0.6849 & 0.6789 \\
BERT & IO & 0.6830 & 0.6814 \\
BERT & BIO & 0.6852 & 0.6815 \\
ALBERT & IO & 0.6621 & 0.6702 \\
ALBERT & BIO & 0.6679 & 0.6431 \\
Biaffine & - & 0.6731 & 0.6627 \\
\hline
\end{tabular}
    \caption{Test and Dev Results of different models on various tagging scheme}
    \label{tab:test_dev_results}
\end{table}
Our best performing architecture proved to be the sequence tagging system with XLnet transformer trained with \emph{IO} tagging scheme.
Additionally, in Table 6 we show how the LSTM and CRF over the transformer architecture , and our pre-processing step mentioned in Section 7 affect the performance of our best performing architecture.
\begin{table}[ht]
    \centering
\begin{tabular}{r c c}
    \hline
     & \textbf{F1} & $\Delta$ \\
    \hline
    Our Model & 0.6922 & - \\
       - LSTM & 0.6912 & 0.0010 \\
        - CRF & 0.6850 & 0.0072 \\
        - Pre-processing & 0.6759 & 0.1630\\
    \hline
\end{tabular}
    \caption{Impact of LSTM, CRF and pre-processing on learning}
    \label{tab:impact_table}
\end{table}

\section{Results Analysis}

\subsection{Length vs Performance}

We wanted to understand how the performance of the system varied with varying lengths of spans. Table 7 summarizes the performance of our best performing systems on all approaches experimented by us, on the test dataset spans, divided into 3 sets according to their length in terms of the number of words that help to make the span.

\begin{table}[ht]
    \centering
    \begin{tabular}{l l l}
    \hline
    Model & Span length & F1 \\
    \hline
    & 1     & \textbf{0.6546}  \\
    Seq. Tagger (IO) &2-4   & 0.1750 \\
    &$>$=5 & 0.0596 \\
    \hline
    & 1     & \textbf{0.6588}  \\
    Seq. Tagger (BIO) &2-4   & 0.1524 \\
    &$>$=5 & 0.09198 \\
    \hline
    & 1     & \textbf{0.6486}  \\
     Dependency Parsing & 2-4   & 0.0514 \\
    &$>$=5 & 0.0 \\
    \hline
    \end{tabular}
    \caption{Span Length vs. Performance}
    \label{tab:len_vs_perform}
\end{table}

\subsection{Learning context}
Majority of single word spans in the dataset are the most commonly used cuss words or abusive words in the English language, i.e., words that can be directly classified as toxic and are not context-dependant, e.g. \emph{"stupid"},\emph{"idiot"} etc., with spans longer than a single word having a lesser ratio of such words. We acknowledge the fact that an AI-based system should be able to do much more, like learning the context behind which a word is used, than just detect common \emph{English} cuss words from a sentence, which can be otherwise done by a simple dictionary search. The deteriorating performance of the model with an increase in span length makes us dig deeper into our test set results to find out if our model is being able to detect \emph{context-based} toxic spans from sentences. We follow a two step procedure to analyze this. First, we calculate our model performance on single-word spans consisting of just the top 25 most commonly occurring context-independent cuss words\footnote{List of cuss words used for analysis can be found in our GitHub repository}. Table 8 shows an analysis of these results. Second, we take the word \emph{"black"} and analyze two sentences in our test where the word black was mentioned in a toxic and non-toxic context. Fig. 5 shows how our model indeed tags the latter black as toxic and the former one as non-toxic.

\begin{table}[ht]
    \centering
    \begin{tabular}{c c}
    \hline
       Single Word Cuss Spans  & Others \\
    \hline
        0.6894  &  0.1736    \\
    \hline
    \end{tabular}
    \caption{F\textsubscript{1} score of context independent cuss words}
    \label{tab:f1_context}
\end{table}

%For analyzing follow a 2-step pwe take the word \emph{"black"} and analyze two sentences in our test set where the word \emph{"black"} was mentioned in a toxic and non-toxic context. Fig 4 shows how our model indeed predicts the utterance \emph{"black"} in the first sentence as toxic, and the latter as not.

%We follow a two step procedure to analyze this.First, we calculate our model performance on spans which do-not contain any context-independent cuss words(star).Second,we take the word \emph{"black"} and analyze two sentences in our test where the word black was mentioned in a toxic and non-toxic context. Fig 3 shows our model is indeed tags the later black as toxic and the first one as non-toxic.

\begin{figure}
    \centering
    \includegraphics[width=0.48\textwidth]{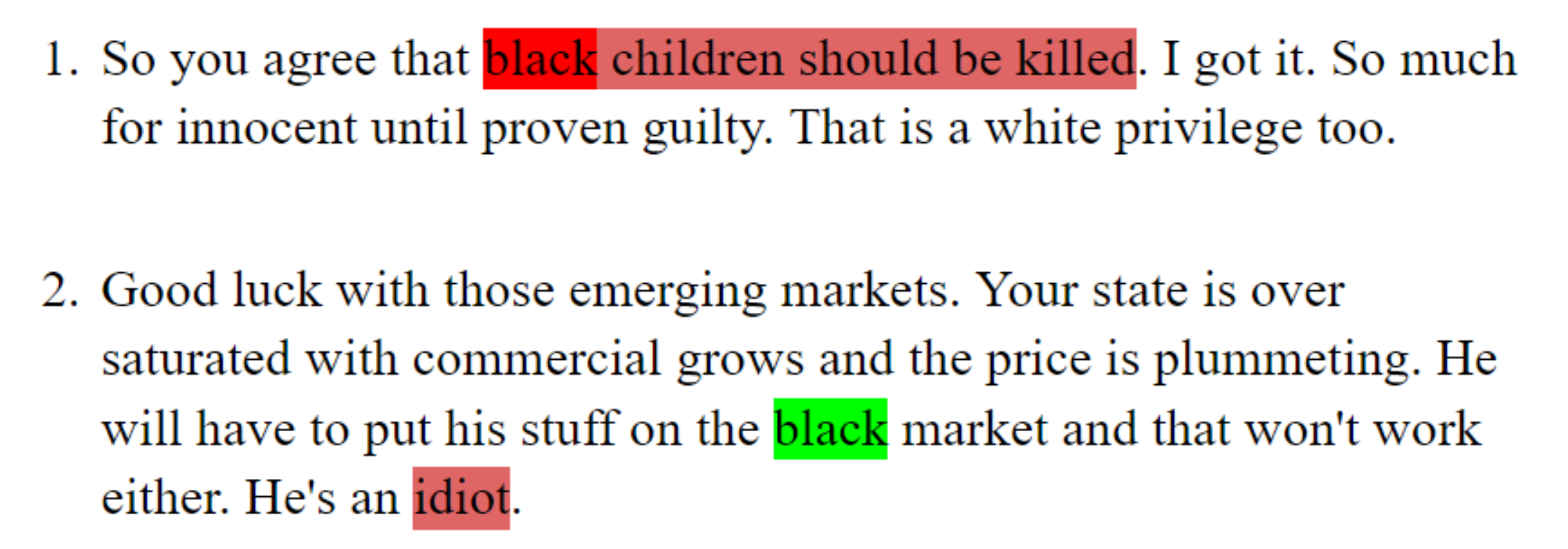}
    \caption{Toxicity classification of the word "black" in toxic and non-toxic context}
    \label{fig:context_img}
\end{figure}

%\subsection{Randomized}

\section{Conclusion}
In this paper, we present our approach to SemEval-2021 Task 5: Toxic Spans Detection. Our best submission gave us an \emph{F\textsubscript{1}} score of \textbf{0.6922}, placing us $7^{th}$ on the Evaluation Phase Leaderboard. Future work includes independently incorporating both post level and sentence level context for determining the toxicity of a word, and also collating a dataset with toxic spans comprising of a healthy mixture of simple cuss words (which can always be attributed as toxic independant of the context) and words for which the toxicity of the word depends on the context in which it appears, thereby making better systems towards \emph{contextual} toxic span detection.

%\section*{Acknowledgments}

\bibliographystyle{acl_natbib}
\bibliography{anthology,acl2021}

%\appendix

\end{document}